\documentclass{article}

    \PassOptionsToPackage{numbers, compress}{natbib}
    \usepackage[preprint]{neurips_2024}

\usepackage[utf8]{inputenc} % allow utf-8 input
\usepackage[T1]{fontenc}    % use 8-bit T1 fonts
\usepackage{hyperref}       % hyperlinks
\usepackage{url}            % simple URL typesetting
\usepackage{booktabs}       % professional-quality tables
\usepackage{amsfonts}       % blackboard math symbols
\usepackage{nicefrac}       % compact symbols for 1/2, etc.
\usepackage{microtype}      % microtypography
\usepackage{xcolor}         % colors
\usepackage{graphicx}
\usepackage{subfig}
\usepackage{enumitem}
\usepackage{bm}
\usepackage{multirow}
\usepackage{wrapfig}
\usepackage{algorithm,algorithmic}
\usepackage{caption}

\title{Adaptive Teaching with Shared Classifier for Knowledge Distillation}

\author{
  Jaeyeon Jang$^{1,2}$\thanks{Corresponding author: \texttt{jaeyeon.jang@catholic.ac.kr}} \\
  \And
  Young-Ik Kim$^{2}$\\
  \And
  Jisu Lim$^{2}$\\
  \And
  Hyeonseong Lee$^{2}$\\
  \And
  \textnormal{$^1$ Department of Data Science, The Catholic University of Korea} \\
  \textnormal{$^2$ AI Lab, DenComm} \\
}

\begin{document}

\maketitle

\begin{abstract}
Knowledge distillation (KD) is a technique used to transfer knowledge from an overparameterized teacher network to a less-parameterized student network, thereby minimizing the incurred performance loss. KD methods can be categorized into offline and online approaches. Offline KD leverages a powerful pretrained teacher network, while online KD allows the teacher network to be adjusted dynamically to enhance the learning effectiveness of the student network. Recently, it has been discovered that sharing the classifier of the teacher network can significantly boost the performance of the student network with only a minimal increase in the number of network parameters. Building on these insights, we propose adaptive teaching with a shared classifier (ATSC). In ATSC, the pretrained teacher network self-adjusts to better align with the learning needs of the student network based on its capabilities, and the student network benefits from the shared classifier, enhancing its performance. Additionally, we extend ATSC to environments with multiple teachers. We conduct extensive experiments, demonstrating the effectiveness of the proposed KD method. Our approach achieves state-of-the-art results on the CIFAR-100 and ImageNet datasets in both single-teacher and multiteacher scenarios, with only a modest increase in the number of required model parameters. The source code is publicly available at \url{https://github.com/random2314235/ATSC}.
\end{abstract}

\section{Introduction}
In recent decades, deep neural networks (DNNs) have achieved significant success across various real-world tasks. These include a range of visual \cite{he_deep_2016, chen_encoder-decoder_2018, ren_faster_2015}, natural language processing \cite{brown_language_2020}, and automatic speech recognition tasks \cite{radford_robust_2023}. However, the achievements of DNNs largely depend on a considerable number of parameters, making them challenging to deploy on resource-constrained edge devices due to their high computational and storage demands. To address this issue, the technique of knowledge distillation (KD) has been employed to distill knowledge from an overparameterized teacher network to a less-parameterized student network \cite{ba_deep_2014, hinton_distilling_2014}. KD aims to significantly enhance the performance of the student model over that attained when the student model is trained independently.

\begin{figure}[h]\centering
\setcounter{subfigure}{-1}
\subfloat{\includegraphics[width=0.7\linewidth]{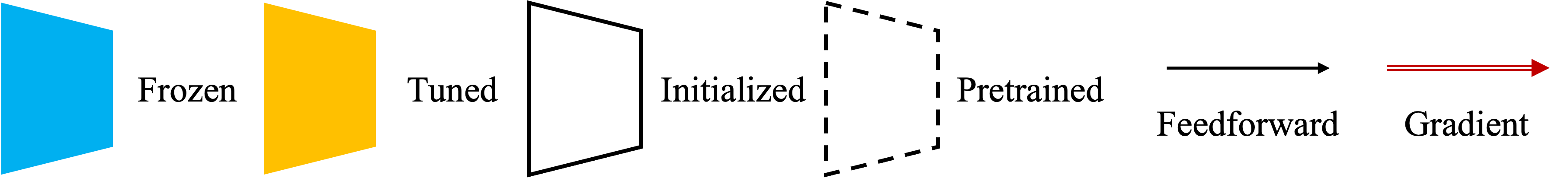}} 
\par
\subfloat[Vanilla KD \label{fig:model_comparison_a}]{\includegraphics[width=0.33\linewidth]{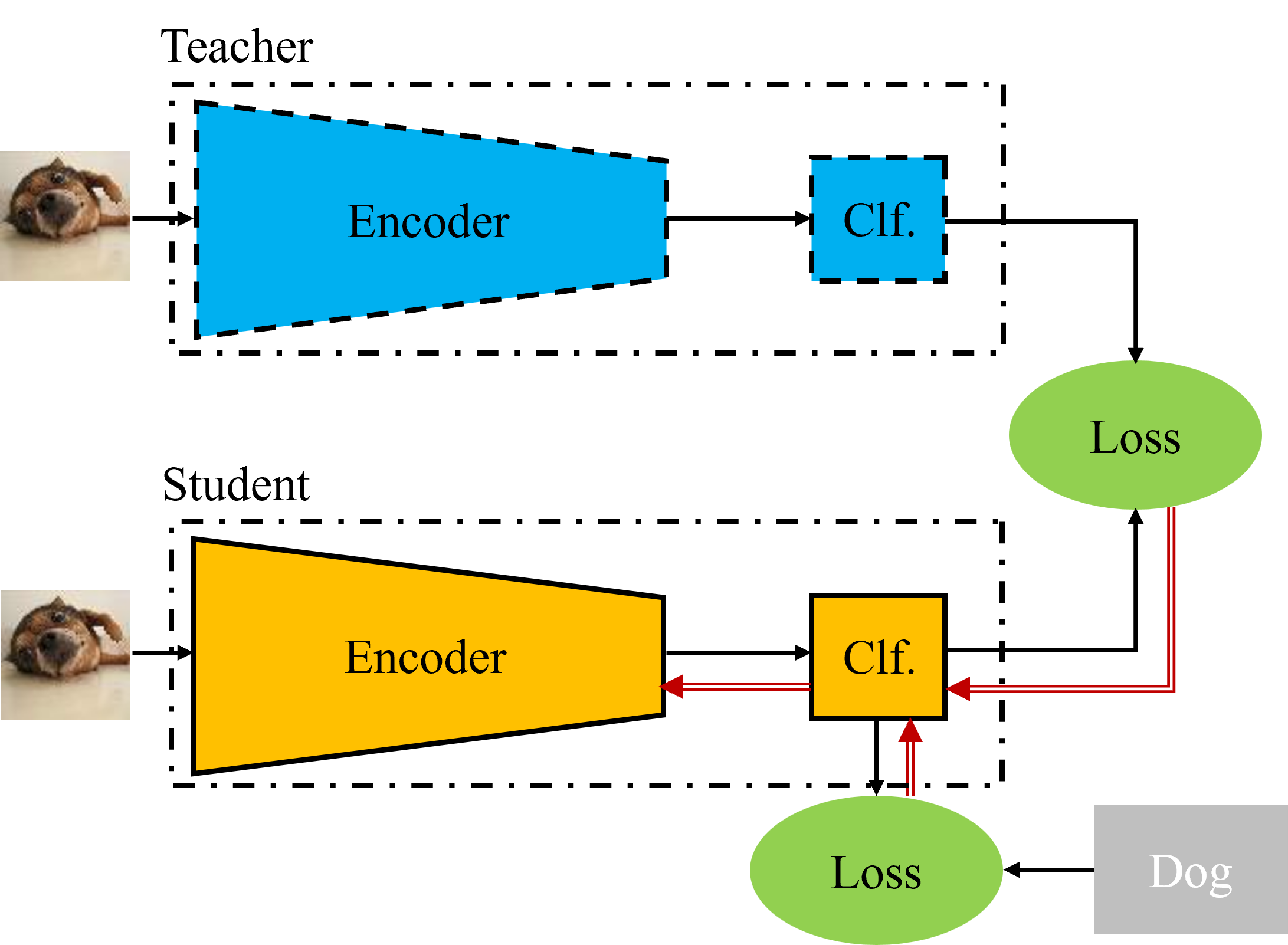}}
\subfloat[Feature distillation \label{fig:model_comparison_b}]{\includegraphics[width=0.33\linewidth]{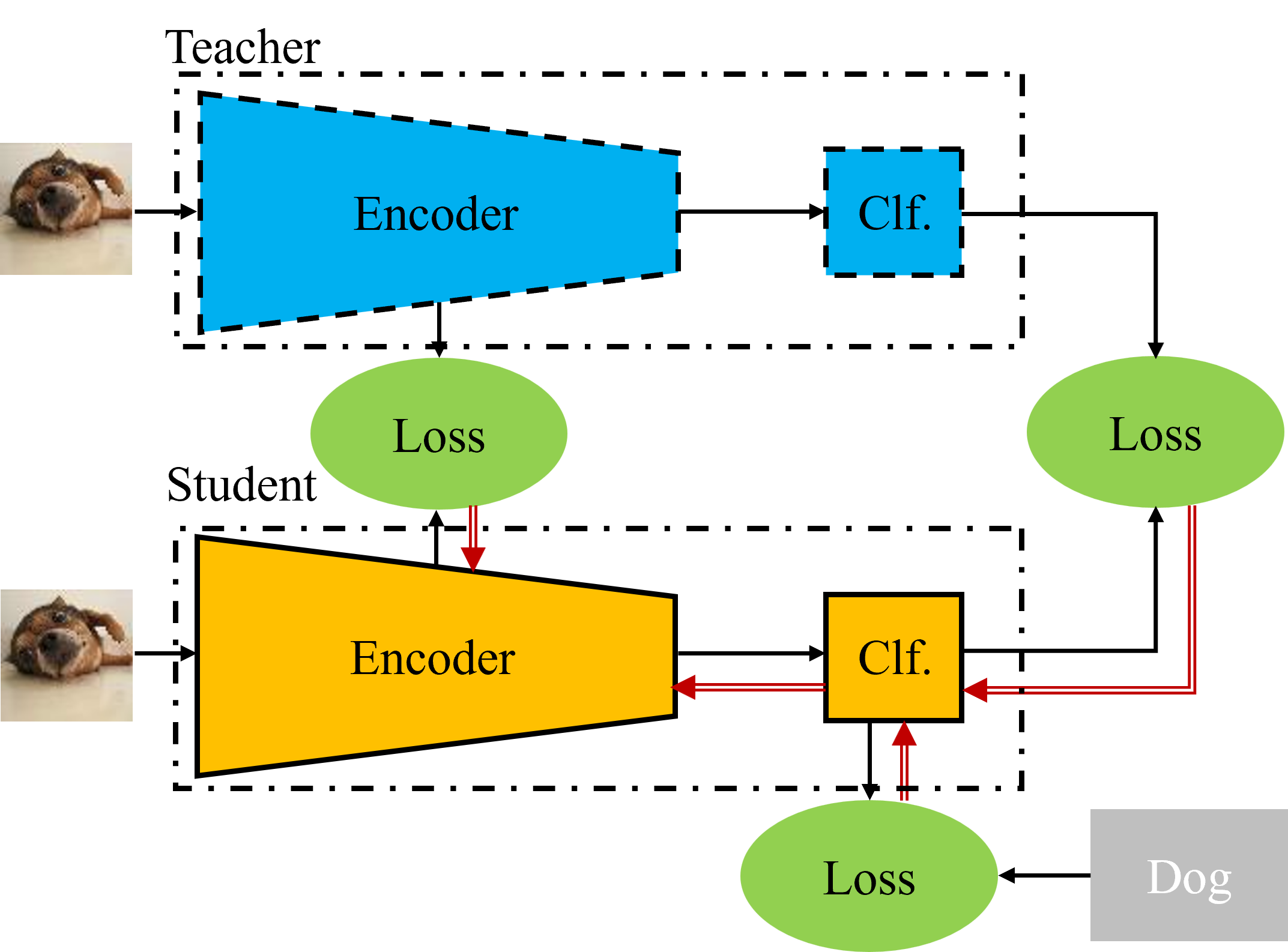}}
\subfloat[Online KD \label{fig:model_comparison_c}]{\includegraphics[width=0.33\linewidth]{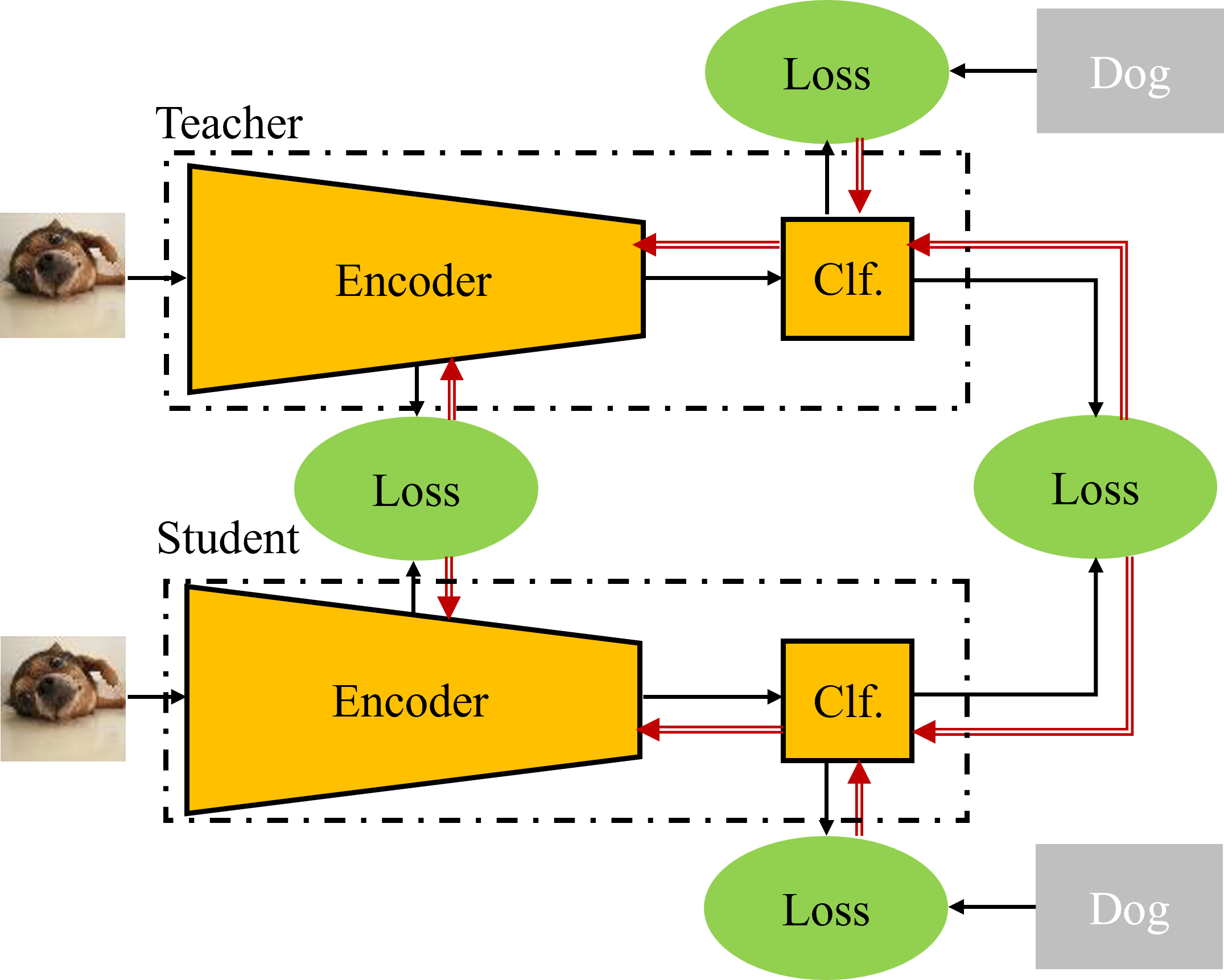}}
\par
\subfloat[SimKD \label{fig:model_comparison_d}]{\includegraphics[width=0.4\linewidth]{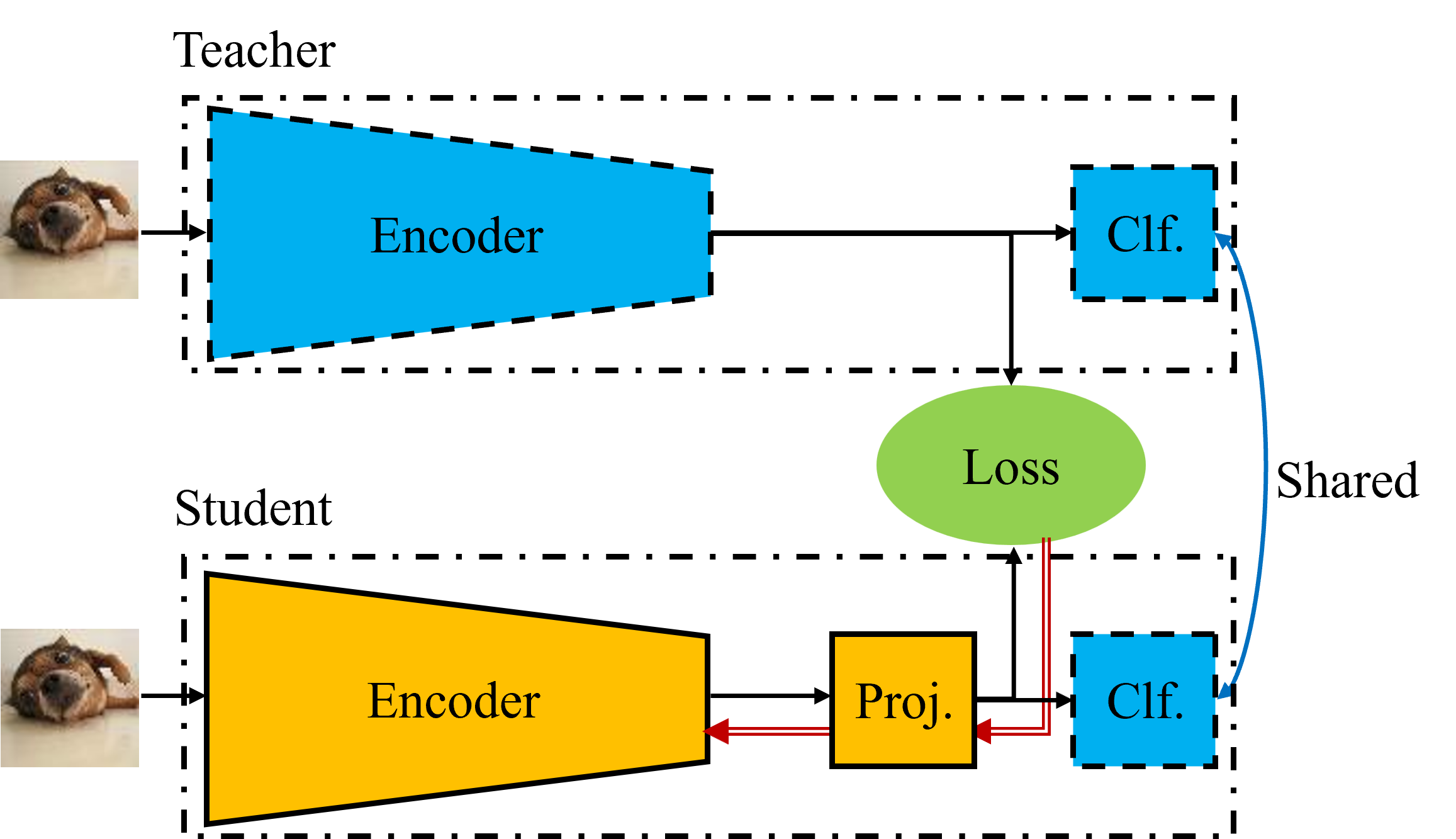}}
\subfloat[ATSC \label{fig:model_comparison_e}]{\includegraphics[width=0.4\linewidth]{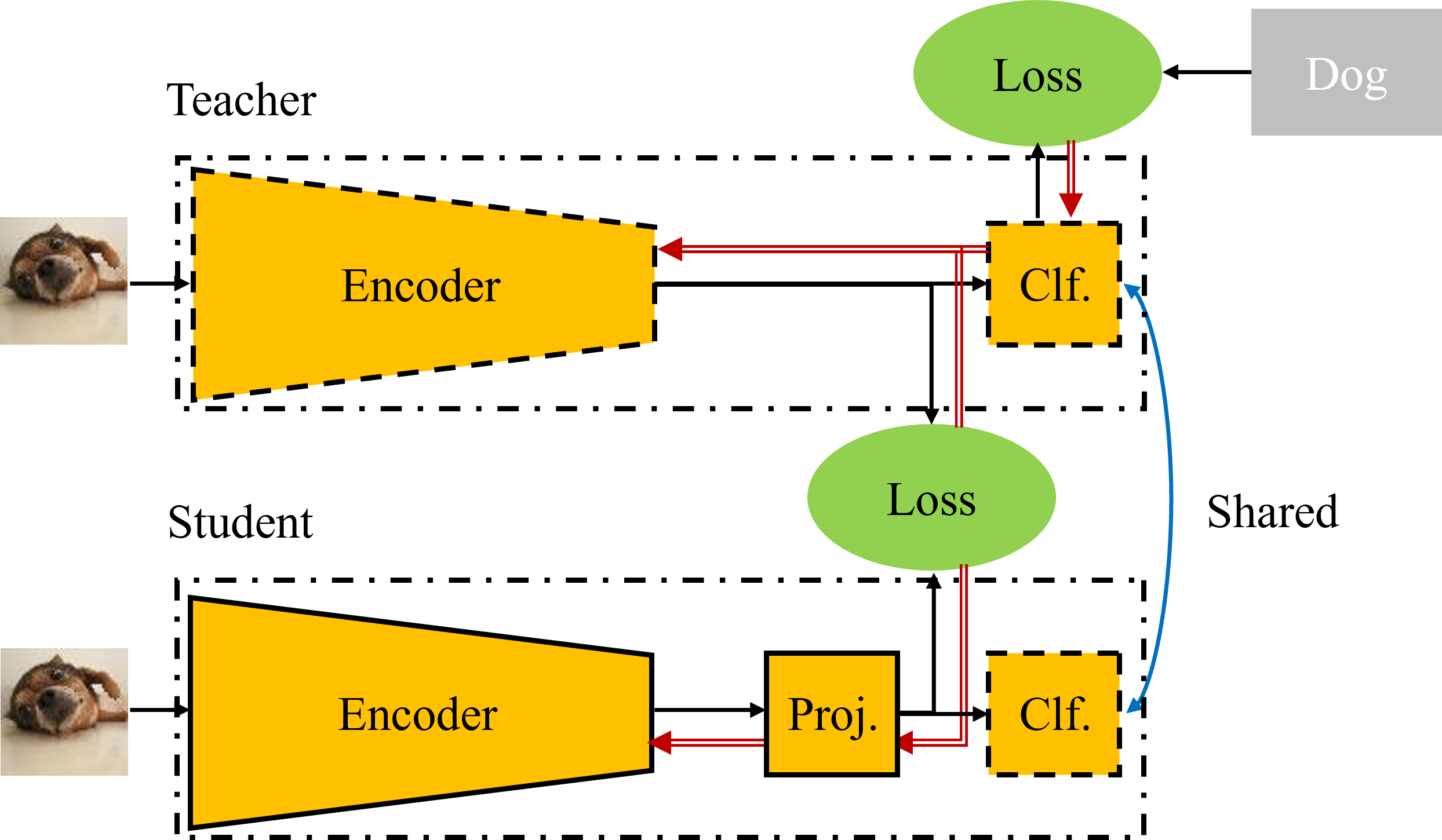}}
\caption{Illustrative comparison among different KD methods. Clf. and Proj. denote a classifier and a projector, respectively. The main differences among these methods include their loss definitions, the flow of gradients, and teacher roles during the learning process. (a) In vanilla KD, gradients are derived from two losses: a loss comparing the last-layer logits of the pretrained teacher and the student and a prediction loss. (b) Feature distillation extends beyond vanilla KD by also extracting gradient information from the intermediate layers of the encoder. (c) In general, online KD is a dynamic form of feature distillation in which both the teacher and the student alternately apply distillation techniques to each other. (d) In SimKD, the student is trained to map its representations to those produced by the encoder of the pretrained teacher; this step is facilitated by an additional projector. This method also involves sharing the classifier of the large teacher network with the smaller student network to maintain high discriminative capabilities. (e) Our proposed ATSC approach enables the teacher to not only guide the student but also adaptively fine-tune its encoder parameters to better support the learning procedure of the student. Furthermore, the classifier is optimized to consider the updated encoder of the teacher, ensuring a more effective and integrated learning process.}
  \label{fig:model_comparison}
  \vspace{-10pt}
\end{figure}

The KD process is often focused on the final layer of the utilized network, as shown in Fig.~\ref{fig:model_comparison_a}, where the student learns ``softened'' versions of the outputs yielded by the teacher network. This approach enables the student to gain insights into the probability landscape that the teacher perceives rather than only hard labels \cite{ba_deep_2014, hinton_distilling_2014}. Two categories of methods are primarily used to apply KD. Offline KD focuses on transferring knowledge from an already-trained, larger teacher model to the student (see Figs.~\ref{fig:model_comparison_a} and \ref{fig:model_comparison_b}). This method allows for a more focused distillation process, leveraging the fully developed knowledge of the teacher model \cite{liu_norm_2023}. Conversely, in the online KD approach shown in Fig.~\ref{fig:model_comparison_c}, both the teacher and student models are trained simultaneously from scratch. This approach allows the student to learn directly from the ongoing learning process of the teacher \cite{guo_online_2020}. However, irrespective of the chosen KD approach, researchers have established the notion that the distillation of intermediate feature maps, often in conjunction with the final layer, yields significantly enhanced effectiveness \cite{liu_norm_2023, tung_similarity-preserving_2019-1, ahn_variational_2019, tian_contrastive_2020, yang_knowledge_2021, chen_cross-layer_2021, li_shadow_2022} (see Figs.~\ref{fig:model_comparison_b} and \ref{fig:model_comparison_c}). Chen \textit{et al.} \cite{chen_knowledge_2022} further enhanced this domain by arguing that the robust predictive ability of the teacher model stems not only from its expressive features but also significantly from its discriminative classifier. They suggested that considerable enhancements can be attained by simply replicating the classifier of the teacher in the student after aligning the features of the student with those of the teacher using a projector that requires only a small number of parameters, as shown in Fig.~\ref{fig:model_comparison_d}.

By reviewing the literature, we identified three critical insights. First, offline KD methods leverage pretrained teacher models that possess strong discriminative capabilities. Conversely, in online KD approaches, the teacher model can be adjusted to enhance the learning effectiveness of the student model. Third, by adopting the classifier of the teacher, the student can significantly enhance its discriminative ability. Integrating these insights, we propose a novel KD method named adaptive teaching with a shared classifier (ATSC), as illustrated in Fig.~\ref{fig:model_comparison_e}. Our approach is characterized as follows. Initially, a teacher network is pretrained to harness the discriminative power of large models. Subsequently, both the teacher and student are trained collaboratively to boost the performance of the student. Specifically, the teacher network adjusts itself to better align with the learning needs of the student based on its capabilities. Essentially, our method leverages the strengths of the online KD approach, starting with a pretrained teacher network. Furthermore, the student gains direct access to the powerful classifier of the teacher, which is enabled by a projector that requires only a small number of parameters.

We conduct extensive experiments on standard benchmark datasets to validate the effectiveness of our proposed ATSC method. The results demonstrate that ATSC consistently outperforms the existing KD methods across various settings. Specifically, on the CIFAR-100 dataset, ATSC achieves a 5.30\% accuracy improvement over the baseline student network without KD in a single-teacher setting and a 6.70\% improvement in a multiple-teacher setting. These results establish ATSC as the new state-of-the-art approach for both settings. Furthermore, on the challenging ImageNet dataset, ATSC enhances the accuracy of the student model (ResNet-18) by 1.19\% with ResNet-50 as the teacher, achieving both the best performance and the fastest training convergence process.

Our contributions are summarized as follows.
\begin{itemize}[leftmargin=1em]
\vspace{-3pt}
\item We believe this study is the first to effectively integrate three key components of KD techniques: a large pretrained teacher network with powerful discriminative capabilities, an adaptive teaching-based KD method that guides the teacher to self-adjust its parameters to enhance the student learning process, and a shared classifier that leverages the capabilities of the teacher.
\vspace{-3pt}
\item The concept of adaptive teaching, initially introduced in the field of KD, involves the teacher model sacrificing a portion of its own discriminative power to more effectively assist the student model in learning representations. The obtained experimental results demonstrate that this slight reduction in the discriminative power of the teacher can lead to significant performance gains for the student model.
\vspace{-3pt}
\item We achieve state-of-the-art performance on the CIFAR-100 and ImageNet datasets under various experimental settings.
\vspace{-3pt}
\item Furthermore, we empirically demonstrate that ATSC is robust across a diverse range of balancing parameter settings required for training, thereby reducing the effort needed for hyperparameter optimization purposes.
\end{itemize}

\section{Related works} \label{sec:RW}
\noindent\textbf{Offline KD approaches}. Vanilla KD was proposed to distill knowledge based on the logits of a teacher network using temperature-scaled soft supervision in an offline manner \cite{hinton_distilling_2014}. Romero \textit{et al.} \cite{romero_fitnets_2015} demonstrated that the intermediate representations learned by the teacher can be utilized as hints to achieve improved distillation performance. Following this work, in the last few years, many follow-up studies have aimed to enhance the transfer of knowledge from a pretrained teacher network by applying diverse techniques, such as feature encoding \cite{yim_gift_2017, srinivas_knowledge_2018}, sample relations encoded using pairwise similarity matrices \cite{passalis_learning_2018, park_relational_2019, tung_similarity-preserving_2019-1} or modeled using contrastive learning \cite{tian_contrastive_2020, xu_knowledge_2020, zhou_distilling_2021}, distribution learning \cite{malinin_ensemble_2020, yang_knowledge_2021}, attention rephrasing \cite{kim_paraphrasing_2018, ji_show_2021}, cross-layer association learning \cite{chen_cross-layer_2021, chen_distilling_2021}, many-to-one representation matching \cite{liu_norm_2023}, and reuse of the classifier possessed by the teacher \cite{chen_knowledge_2022}. Search methods based on reinforcement learning have also recently been proposed to improve the feature distillation process \cite{liu_search_2020, yue_matching_2020, deng_distpro_2022}.

\noindent\textbf{Online KD approaches}. This category of approaches has been less studied because they require more training time than offline approaches. Nonetheless, the principle of tailoring instructions to the aptitude of the student network to maximize its potential should not be overlooked \cite{li_shadow_2022}. This category focuses on jointly training multiple models. For example, Zhang \textit{et al.} \cite{zhang_deep_2018} used mutual learning to jointly train a set of models from scratch, where each model acted as a teacher to the others. Lan \textit{et al.} \cite{lan_knowledge_2018} introduced a strategy to guide the training processes of individual branches (students) using a multibranch ensemble (teacher). Similarly, Wu \textit{et al.} \cite{wu_peer_2021} enhanced the collaboration among peers through mutual peer distillation. Additionally, Guo \textit{et al.} \cite{guo_online_2020} implemented a dynamic ensemble of soft predictions derived from multiple branches, distorting the input samples to create soft targets for branch supervision. Recently, researchers have explored the diversity in the logits of branches through feature fusion and learning in combination with classifier diversification \cite{kim_feature_2020, li_online_2020}. Additionally, several effective strategies have been employed to enhance online KD: feature-level adversarial training \cite{chung_feature-map-level_2020}, two-level distillation based on an attention mechanism \cite{chen_online_2020}, and gradual hierarchical distillation \cite{gong_adaptive_2023}.

\section{Method}
\subsection{Background: KD with a reused teacher classifier}
In this section, we revisit the concept of KD with a reused teacher classifier \cite{chen_knowledge_2022}. The core premise of using a pretrained teacher classifier is based on the assumption that the given data contain capability-invariant information that can be easily transferred between different models. Additionally, the final classifier of the teacher model often holds crucial capability-specific information, which may be challenging for a simpler student model to replicate. Therefore, during the learning process, this method focuses on transferring knowledge derived solely from the output of the encoder, which is directly fed into the classifier.

Let $\bm{x}$ be an input sample and $\bm{y}$ be its corresponding ground-truth label. Consider $E_T$ and $E_S$ as the encoders of a pretrained teacher network and a student network, respectively, and let $C$ be their shared classifier. The corresponding parameter sets are $\bm{\theta}_{E_T}$, $\bm{\theta}_{E_S}$, and $\bm{\theta}_{C}$. Then, the objective for this method is formulated as
\begin{equation} \label{eq_simKD}
\min_{\bm{\theta}_{E_S}, \bm{\theta}_{\mathcal{P}}} \mathcal{L}_{MSE}(E_T(\bm{x}), \mathcal{P}(E_S(\bm{x}))),
\end{equation}
where $\mathcal{L}_{MSE}$ is the mean squared error (MSE) loss function and $\mathcal{P}$ represents a projector, which is introduced to match the feature dimensions of the outputs obtained from both encoders with the corresponding parameter set $\bm{\theta}_{\mathcal{P}}$. It has been shown that the introduction of a projector can significantly alleviate the performance degradation incurred from the teacher to the student, with a relatively small increase in the number of required parameters. Detailed information about the architecture of the projector is available in our Appendix.

\subsection{Adaptive teaching with a shared classifier}
\begin{figure}[h]\centering
\subfloat[Step 1 \label{fig:method_1}]{\includegraphics[width=0.38\linewidth]{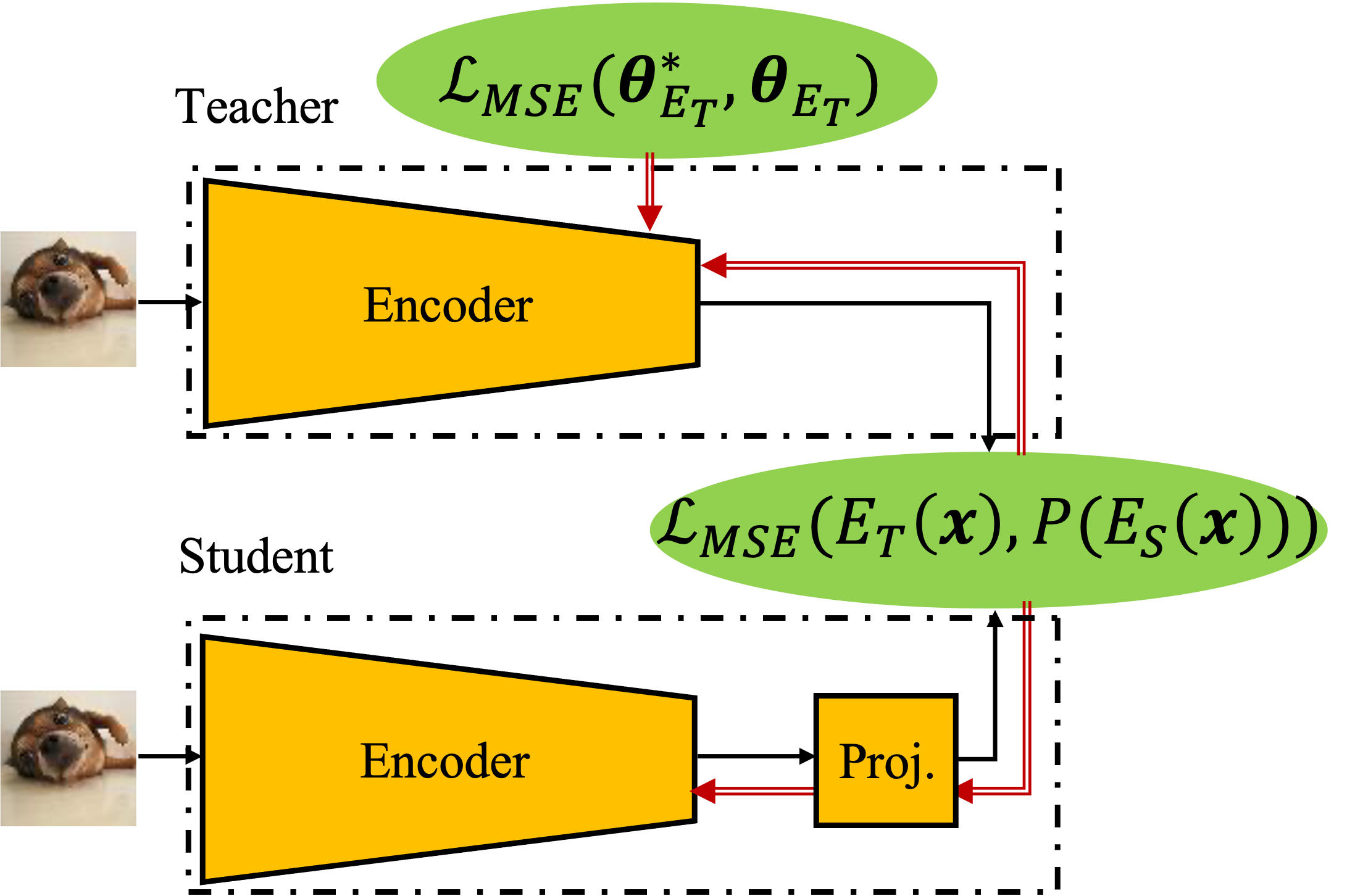}}
    \begin{minipage}[b]{0.38\textwidth}
\subfloat{\includegraphics[width=\linewidth]{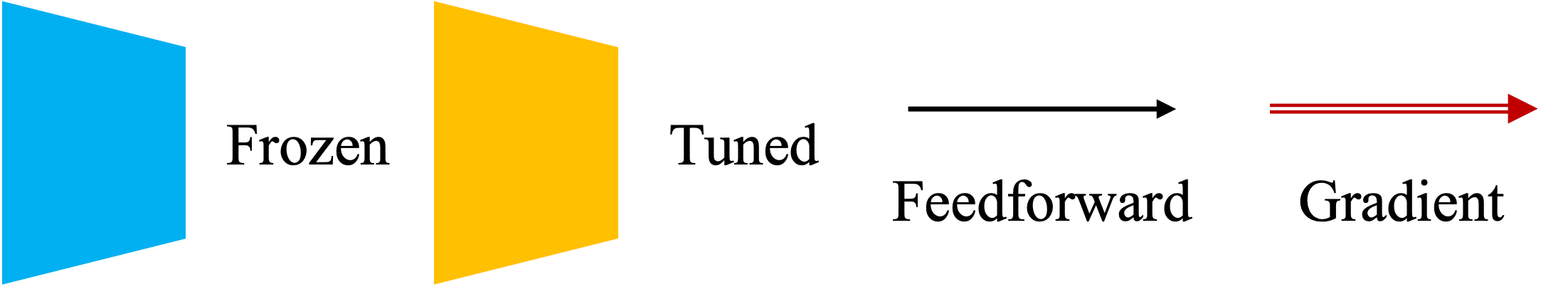}}\\[20pt]
\setcounter{subfigure}{1}
\subfloat[Step 2 \label{fig:method_2}]{\includegraphics[width=\linewidth]{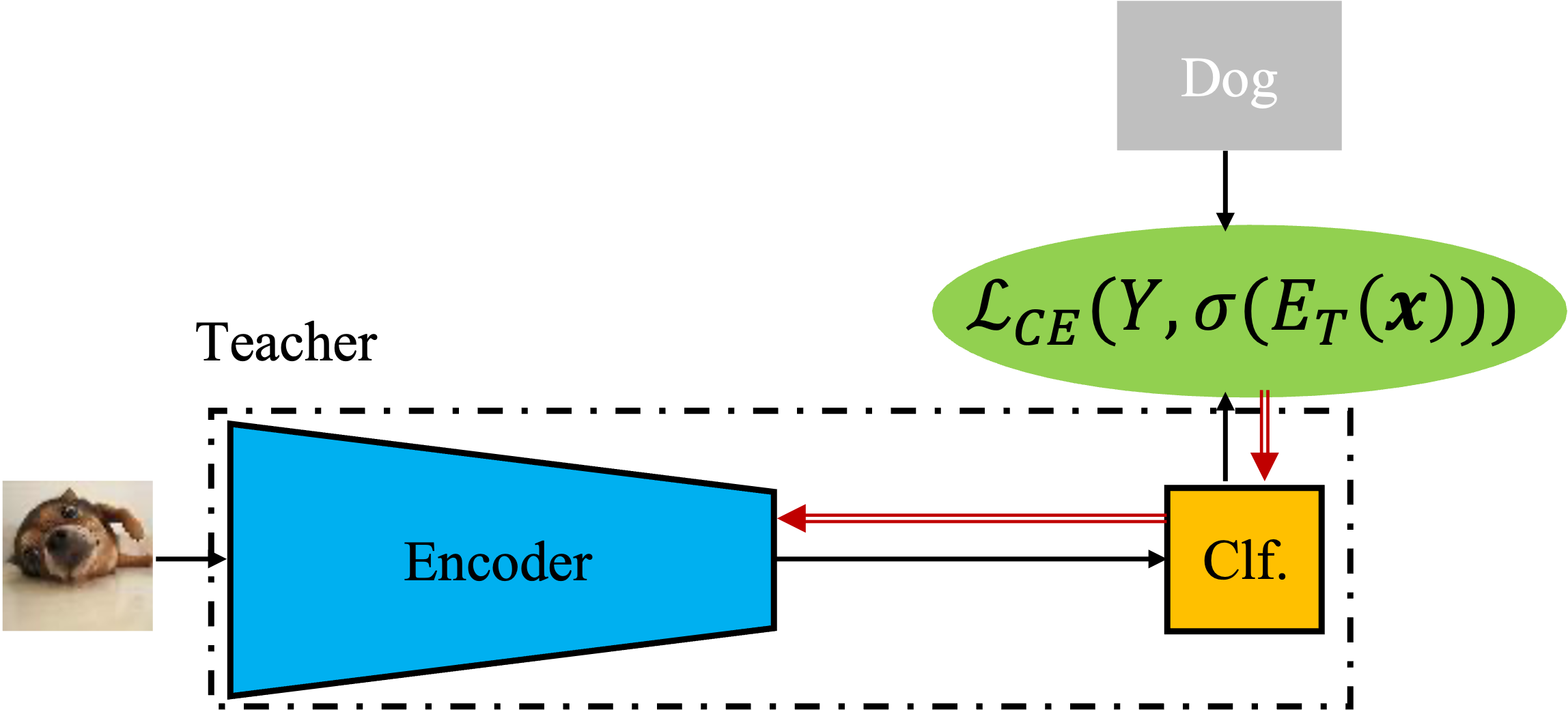}}
\vfill
    \end{minipage}
\caption{An overview of the proposed ATSC approach. In (a), $\mathcal{L}_{MSE}(E_T(\bm{x}), \mathcal{P}(E_S(\bm{x})))$ represents the MSE loss between the representations derived from the teacher and student models, and $\mathcal{L}_{MSE}(\bm{\theta}^*_{E_T}, \bm{\theta}_{E_T})$ denotes the penalty imposed on the parameter changes exhibited by the encoder of the teacher. In (b), $\mathcal{L}_{CE}(\bm{y}, \sigma(C(E_T(\bm{x}))))$ denotes the cross-entropy loss.}
  \label{fig:method}
\end{figure}

In this section, we propose a novel KD method comprising two steps, as shown in Fig. \ref{fig:method}. If the capability-specific information of the teacher model is not readily captured by a simpler student model, the teacher should adjust to provide information that is more easily learnable, even though it may not be optimal from the perspective of the teacher. Therefore, we allow some distortion from the pretrained teacher network if it enables the student model to more effectively match its representations. With this objective, we first optimize the encoders of both networks based on the following objective function:
\begin{equation} \label{eq_step1}
\min_{\bm{\theta}_{E_T}, \bm{\theta}_{E_S}, \bm{\theta}_{\mathcal{P}}} \mathcal{L}_{MSE}(E_T(\bm{x}), \mathcal{P}(E_S(\bm{x})))+\alpha * \mathcal{L}_{MSE}(\bm{\theta}^*_{E_T}, \bm{\theta}_{E_T}),
\end{equation}
where $\bm{\theta}^*_{E_T}$ represents the parameter set of the pretrained teacher encoder before conducting collaborative learning between the teacher and student models, and $\alpha$ is a balancing parameter that constrains the distortion within $\bm{\theta}_{E_T}$. Through this process, the student model can acquire more knowledge by learning representations that are easier for its encoder to produce.

However, this process may inevitably degrade the classifier of the teacher since it relies on the undistorted outputs of the teacher encoder. To maximally maintain the discriminative power of the classifier, it should be fine-tuned following changes in its input space. Thus, we update the shared classifier based on the following equation:
\begin{equation} \label{eq_step2}
\min_{\bm{\theta}_{C}} \mathcal{L}_{CE}(\bm{y}, \sigma(C(E_T(\bm{x})))),
\end{equation}
where $\mathcal{L}_{CE}$ denotes the standard cross-entropy loss and $\sigma$ represents the softmax function. In this second step, the encoder of the teacher remains frozen. During the training process, equations (\ref{eq_step1}) and (\ref{eq_step2}) are alternated for each batch. The complete pseudocode of our ATSC method can be found in the Appendix.

The shared classifier can be updated with the projector based on the encoder of the student by using the loss function $\mathcal{L}_{CE}(\bm{y}, \sigma(C(\mathcal{P}(E_S(\bm{x})))))$. However, we argue that leveraging the classifier of the teacher model based on its encoder is crucial for minimizing the performance loss incurred by the student model. Related experimental results are detailed in Section \ref{sec:ablation}.

\subsection{Extending ATSC to multiteacher models}
Our proposed ATSC method is readily adaptable to scenarios in which multiple teachers are available for the student training process. Let $T_1, \cdots, T_N$ represent $N$ teachers. Then, (\ref{eq_step1}) is extended to the following:
\begin{equation} \label{eq_step1_mult}
\min_{\{\bm{\theta}_{E_{T_i}}|i \in \{1, \cdots, N\}\}, \bm{\theta}_{E_S}, \{\bm{\theta}_{\mathcal{P}_i}|i \in \{1, \cdots, N\}\}} \sum_{i=1}^N \mathcal{L}_{MSE}(E_{T_i}(\bm{x}), \mathcal{P}_i(E_S(\bm{x})))+\alpha * \mathcal{L}_{MSE}(\bm{\theta}^*_{E_{T_i}}, \bm{\theta}_{E_{T_i}}),
\end{equation}
where $P_i$ denotes the projector for teacher $i$. In this first step, the student model learns from the average adjusted representations of the teachers. Subsequently, we fine-tune the classifiers to ensure that the average teacher outputs accurately map to the ground-truth labels as follows:
\begin{equation} \label{eq_step2_mult}
\min_{\{\bm{\theta}_{C_i}|i \in \{1, \cdots, N\}\}} \mathcal{L}_{CE}(\bm{y}, \sigma(\frac{1}{N}\sum_{i=1}^N C_i(E_{T_i}(\bm{x})))),
\end{equation}
where $C_i$ denotes the classifier of teacher $i$. Finally, by using the optimized projectors and shared classifiers, the student predicts the ground-truth label by applying the equation below:
\begin{equation} \label{eq_step2_mult}
\sigma(\frac{1}{N}\sum_{i=1}^N C_i(P_i(E_S(\bm{x})))).
\end{equation}

\section{Experiments} \label{sec:experiments}
% Follow the structure of "Adaptive Hierarchy-Branch Fusion for Online Knowledge Distillation"
In this section, we conduct extensive experiments to validate the effectiveness of our proposed ATSC method on two benchmark datasets: CIFAR-100 \cite{krizhevsky_learning_2009} and ImageNet \cite{russakovsky_imagenet_2015}. We initially compare the performance of ATSC with that of diverse state-of-the-art offline and online KD methods in scenarios involving both single-teacher and multiteacher configurations. Subsequently, we conduct ablation studies and sensitivity analyses to further substantiate the contributions of our approach.

%do we include sensitivity analysis here or not. According to it, check.

\subsection{Experimental setups}
\textbf{Training details.}
We follow the training procedure used in prior studies \cite{tian_contrastive_2020, chen_cross-layer_2021, chen_knowledge_2022}. Specifically, we deploy the stochastic gradient descent optimizer with a Nesterov momentum of 0.9 across all datasets. We use a batch size of 64 and apply weight decay rates of $5 \times 10^{-4}$ for CIFAR-100 and $1 \times 10^{-4}$ for ImageNet. For CIFAR-100, the training process spans 240 epochs, with the learning rate reduced by a factor of 10 at the 150th, 180th, and 210th epochs. The starting learning rates are 0.01 for the models in the MobileNet/ShuffleNet series and 0.05 for the other models. The balancing parameter $\alpha$ is typically set to 1, and the reduction factor, which influences the number of filters in the convolutional layers of the projectors, is set to 2 by default, as recommended by Chen \textit{et al.} \cite{chen_knowledge_2022}. We only report hyperparameter settings for the cases in which different values are applied to obtain the experimental results. In the case of ImageNet, training lasts 120 epochs, starting with an initial learning rate of 0.1, which is decreased by a factor of 10 at the 30th, 60th, and 90th epochs. $\alpha$ is set to 10, and the reduction factor is set to 2 for this large-scale dataset. The details of the two utilized datasets can be found in our Appendix.

In this study, all the experiments are conducted using the PyTorch framework \cite{paszke_pytorch_2019}. The models are trained on a machine equipped with an Intel i9-12900k CPU and two NVIDIA GeForce RTX 4090 GPUs, each with 24 GB of RAM. However, only one GPU is utilized for all the experiments.

\subsection{Comparison with the state-of-the-art KD methods}
\textbf{Results on CIFAR-100.} We compare various KD methods with our approach across a range of teacher-student combinations using popular network architectures. The compared KD methods and the details of the utilized network architectures are both summarized in our Appendix. We conduct comparison experiments exclusively in scenarios, among those used in SimKD~\cite{chen_knowledge_2022} and SHAKE~\cite{li_shadow_2022}, where the accuracy gaps between the teacher model and the student model trained with the optimal KD method for those scenarios exceed 1\%. Specifically, we focus on environments where significant room remains for achieving performance improvements.

\begin{table}[h] %SimKD + Norm
\vspace{-5pt}
\scriptsize
\centering
\caption{Comparison among the top-1 test accuracies (\%) achieved by various KD methods on CIFAR-100. The results of the other KD methods are obtained from the papers that published SimKD~\cite{chen_knowledge_2022} and NORM~\cite{liu_norm_2023}. To ensure a fair comparison, we present the performance of the teacher network used for ATSC alongside those of the other methods (indicated in parentheses). `Student' refers to the performance attained by the student network trained without any KD method. We report the mean accuracy $\pm$ standard deviation achieved over 4 runs by following the protocol described in the SimKD paper~\cite{chen_knowledge_2022}. For ReviewKD~\cite{chen_distilling_2021}, DistPro~\cite{deng_distpro_2022}, and NORM~\cite{liu_norm_2023}, the mean accuracies produced over 5 runs are reported, as provided in the work of NORM~\cite{liu_norm_2023}. The best result achieved under each setting is highlighted in bold.}
\label{Table_1}
\begin{tabular}{l|cccccc}
\specialrule{2pt}{1pt}{1pt}
Teacher & ResNet-32x4 & ResNet-32x4 & ResNet-32x4 & ResNet-32x4 & WRN-40-2 & ResNet-32x4\\
Student & VGG-8 & ShuffleNetV2 & ShuffleNetV1 & WRN-16-2 & MobileNetV2 & MobileNetV2x2 \\
\hline
Teacher & 79.32 (79.42) & 79.32 (79.42) & 79.32 (79.42) & 79.32 (79.42) & 76.44 (76.31) & 79.32 (79.42)\\
Student & 70.46$\pm$0.29 & 72.60$\pm$0.12 & 71.36$\pm$0.25 & 73.51$\pm$0.32 & 65.43$\pm$0.29 & 69.06$\pm$0.10 \\
\hline
KD~\cite{hinton_distilling_2014} &72.73$\pm$0.15&75.6$\pm$0.21&74.3$\pm$0.16&74.9$\pm$0.29&69.07$\pm$0.47&72.43$\pm$0.32\\
FitNet~\cite{romero_fitnets_2015}&72.91$\pm$0.18&75.82$\pm$0.22&74.52$\pm$0.03&74.7$\pm$0.35&68.64$\pm$0.27&73.09$\pm$0.46\\
AT~\cite{zagoruyko_paying_2017}&71.9$\pm$0.13&75.41$\pm$0.1&75.55$\pm$0.19&75.38$\pm$0.18&68.62$\pm$0.31&73.08$\pm$0.14\\
SP~\cite{tung_similarity-preserving_2019-1}&73.12$\pm$0.1&75.77$\pm$0.08&74.69$\pm$0.32&75.16$\pm$0.32&68.73$\pm$0.17&72.99$\pm$0.27\\
VID~\cite{ahn_variational_2019}&73.19$\pm$0.23&75.22$\pm$0.07&74.76$\pm$0.22&74.85$\pm$0.35&68.91$\pm$0.33&72.7$\pm$0.22\\
CRD~\cite{tian_contrastive_2020}&73.54$\pm$0.19&77.04$\pm$0.61&75.34$\pm$0.24&75.65$\pm$0.08&70.28$\pm$0.24&73.67$\pm$0.26\\
SPRL~\cite{yang_knowledge_2021}&73.23$\pm$0.16&76.19$\pm$0.35&75.18$\pm$0.39&75.46$\pm$0.13&69.34$\pm$0.16&73.48$\pm$0.36\\
SemCKD~\cite{chen_cross-layer_2021}&75.27$\pm$0.13&77.62$\pm$0.32&76.31$\pm$0.2&75.65$\pm$0.23&69.88$\pm$0.30&73.98$\pm$0.32\\
SimKD~\cite{chen_knowledge_2022}&75.76$\pm$0.12&78.39$\pm$0.27&77.18$\pm$0.26&77.17$\pm$0.32&70.71$\pm$0.41&75.43$\pm$0.26\\
ReivewKD~\cite{chen_distilling_2021}&N/A&77.78&77.45&N/A&N/A&N/A\\
DistPro~\cite{deng_distpro_2022}&N/A&77.54&77.18&N/A&N/A&N/A\\
NORM~\cite{liu_norm_2023}&N/A&78.32&\textbf{77.79}&N/A&N/A&N/A\\
\hline
ATSC&\textbf{76.31$\pm$0.39}&\textbf{78.84$\pm$0.13}&77.76$\pm$0.08&\textbf{77.34$\pm$0.15}&\textbf{71.18$\pm$0.33}&\textbf{76.18$\pm$0.14}\\
\specialrule{2pt}{1pt}{1pt}	
\end{tabular}
\vspace{-5pt}
\end{table}

\begin{table}[h] %Shadow + Norm, for simkd.../ should change the last case of ATSC with new result with reduction factor of 4
\scriptsize
\centering
\caption{Comparison among the top-1 test accuracies (\%) achieved by various KD methods on CIFAR-100. We present the performance of the teacher network used for ATSC alongside those of the other methods (indicated in parentheses). We report the mean accuracy attained over 5 runs by following the protocol described in the SHAKE paper~\cite{li_shadow_2022}. The results of the other KD methods are obtained from the papers that published SimKD~\cite{chen_knowledge_2022}, SHAKE~\cite{li_shadow_2022}, and NORM~\cite{liu_norm_2023}. The results of SimKD represent the mean accuracy $\pm$ standard deviation values obtained over 4 runs. The reduction factor of the projector is set to 1 for the ResNet-110 \& and ResNet-20 scenario, 4 for the ResNet-50 \& VGG-8 scenario, and 2 for the other two scenarios. The balancing parameter $\alpha$ is set to 10 for the ResNet-110 \& and ResNet-20 scenario and 1 for the other scenarios.}
\label{Table_2}
\begin{tabular}{l|cccc}
\specialrule{2pt}{1pt}{1pt}								
Teacher & ResNet-110 & ResNet-32x4 & VGG-13 & ResNet-50\\
Student & ResNet-20 & ResNet-8x4 & MobileNetV2 & VGG-8\\
\hline
Teacher & 73.88 (74.31) & 79.32 (79.42) & 74.85 (74.64) & 78.87 (79.34)\\
Student & 69.06 & 72.50 & 64.60 & 70.36\\
\hline
KD~\cite{hinton_distilling_2014} & 70.66 & 73.33 & 67.37 & 73.81\\
FitNet~\cite{romero_fitnets_2015} & 68.99 & 73.50 & 64.14 & 70.69\\
SP~\cite{tung_similarity-preserving_2019-1} & 70.04 & 72.94 & 66.30 & 73.34\\
RKD~\cite{park_relational_2019} & 69.25 & 71.90 & 64.52 & 71.50\\
CRD~\cite{tian_contrastive_2020} & 71.46 & 75.51 & 69.73 & 74.30\\
SRRL~\cite{yang_knowledge_2021} & 70.78 & 75.71 & N/A & N/A\\
ReviewKD~\cite{chen_distilling_2021} & N/A & 75.63 & 70.37 & N/A\\
SimKD~\cite{chen_knowledge_2022} & N/A & 78.08$\pm$0.15 & N/A & N/A\\
T$f$-FD~\cite{li_self-regulated_2022} & 70.62 & 73.62 & N/A & N/A\\
ONE~\cite{lan_knowledge_2018} & 70.77 & N/A & 66.26 & 74.35\\
KDCL~\cite{guo_online_2020} & 70.36 & 74.03 & 65.76 & 73.03\\
MetaDistil~\cite{zhou_bert_2022} & 71.40 & N/A & N/A & 74.42\\
DML~\cite{zhang_deep_2018} & 71.52 & 74.30 & 68.52 & 74.22\\
SHAKE~\cite{li_shadow_2022} & \textbf{72.02$\pm$0.08} & 77.35$\pm$0.28 & 70.03$\pm$0.28 & 74.76$\pm$0.28\\
NORM~\cite{liu_norm_2023} & 72.00 & 76.98 & 69.38 & 75.67\\
\hline
ATSC & 69.28$\pm$0.30 & \textbf{78.27$\pm$0.12} & \textbf{70.21$\pm$0.22} & \textbf{76.53$\pm$0.21}\\
\specialrule{2pt}{1pt}{1pt}	
\end{tabular}
% \vspace{-10pt}
\end{table}						

As shown in Tables \ref{Table_1} and \ref{Table_2}, ATSC provides a 5.30\% accuracy gain to the baseline student model, with a maximum gain of 7.12\%. Overall, ATSC demonstrates highly competitive results by achieving the best performance in 8 scenarios and the second-best performance in 1 scenario out of a total of 10 scenarios. This performance improvement comes at the cost of approximately 6.11\% more parameters used. Considering that the teacher network achieves an 8.10\% performance improvement at the cost of a 78.06\% increase in the number of parameters, the efficiency of ATSC in terms of the ratio of parameter usage to performance gain is substantial. Details regarding the changes in the number of parameters required by the student network are provided in Tables \ref{Table_pram} and \ref{Table_pram2}. We also achieve a 0.44\% average accuracy improvement over SimKD, which uses the same number of parameters for its student. Since the distillation loss between the encoders remains the same, as shown in (\ref{eq_simKD}) and (\ref{eq_step1}), this result underscores the contribution of the self-adaptive teaching process executed by the pretrained teacher.

\begin{table}[h] %SimKD + Norm
\scriptsize
\centering
\caption{Numbers of parameters required by the teacher model, the student model without a projector, and the student model with a projector (used for ATSC) for the scenarios in Table \ref{Table_1}. The ratios of the increases in the number of required parameters caused by adding the projector to the student model relative to the teacher model parameters are also reported.}
\label{Table_pram}
\begin{tabular}{l|cccccc}
\specialrule{2pt}{1pt}{1pt}
Teacher & ResNet-32x4 & ResNet-32x4 & ResNet-32x4 & ResNet-32x4 & WRN-40-2 & ResNet-32x4 \\
Student & VGG-8 & ShuffleNetV2 & ShuffleNetV1 & WRN-16-2 & MobileNetV2 & MobileNetV2x\\
\hline
Teacher & 7.4 M & 7.4 M & 7.4 M & 7.4 M & 2.3 M & 7.4 M\\
Student (w/o projector) & 4.0 M & 1.4 M & 0.9 M & 0.7 M & 0.8 M & 2.4 M\\
Student (w/ projector) & 4.2 M & 1.7 M & 1.3 M & 0.9 M & 1.0 M & 2.7 M\\
\hline
Increase (\%) & 3.66 & 4.55 & 4.44 & 3.00 & 6.23 & 4.99 \\
\specialrule{2pt}{1pt}{1pt}	
\end{tabular}
\vspace{-10pt}
\end{table}

\begin{table}[h] %SimKD + Norm
\scriptsize
\centering
\caption{Numbers of parameters required by the teacher model, the student model without a projector, and the student model with a projector (used for ATSC) for the scenarios in Table \ref{Table_2}.}
\label{Table_pram2}
\begin{tabular}{l|cccccccccc}
\specialrule{2pt}{1pt}{1pt}
Teacher & ResNet-110 & ResNet-32x4 & VGG-13 & ResNet-50\\
Student & ResNet-20 & ResNet-8x4 & MobileNetV2 & VGG-8\\
\hline
Teacher & 1.7 M & 7.4 M & 9.5 M & 23.7 M\\
Student (w/o projector) & 0.3 M & 1.2 M & 0.8 M & 4.0 M \\
Student (w/ projector) & 0.3 M & 1.5 M & 1.9 M & 7.8 M\\
\hline
Increase (\%) & 2.99 & 3.22 & 11.65 & 16.37\\
\specialrule{2pt}{1pt}{1pt}	
\end{tabular}
% \vspace{-10pt}
\end{table}

\textbf{Multiteacher KD results.} We also conduct comparative experiments to demonstrate the applicability of our approach in multiteacher scenarios. As demonstrated in Table \ref{Table_3}, ATSC outperforms all the other scenarios, achieving a 6.70\% accuracy improvement over the baseline student model. It also yields an average improvement of 0.71\% over SimKD despite both models requiring the same number of parameters.

\begin{table}[h]
\vspace{-5pt}
\scriptsize
\centering
\caption{Top-1 test accuracies (\%) achieved on CIFAR-100 across multiteacher scenarios. We report the mean accuracy $\pm$ standard deviation values produced over 4 runs. ShuffleNetV2 is used as the student model. AVEG represents a simplified version of vanilla KD that learns average predictions from multiple teachers. We report the performance achieved by the teachers used for ATSC, with the performance of the teachers used by the other KD methods provided in parentheses, as reported in \cite{chen_knowledge_2022}.}
\label{Table_3}
\begin{tabular}{l|cc}
\specialrule{2pt}{1pt}{1pt}								
\multirow{2}{*}{Teachers} & \multirow{2}{*}{Three ResNet-32x4} & Two ResNet-32x4\\
 & & \& One ResNet-110x2\\
\hline
\multirow{2}{*}{Teachers} & 79.32, 79.64, 79.35 & 79.32, 79.64, 78.52\\
 & (79.32, 79.43, 79.45) & (79.43, 79.45, 78.18)\\
Student & 72.60$\pm$0.12 & 72.60 $\pm$ 0.12\\
\hline
AVEG & 75.94 $\pm$ 0.20 & 76.33 $\pm$ 0.14\\
AEKD~\cite{du_agree_2020} & 75.99 $\pm$ 0.18 & 76.17 $\pm$ 0.43\\
AEKD-F~\cite{du_agree_2020} & 77.24 $\pm$ 0.32 & 77.08 $\pm$ 0.28\\
SimKD~\cite{chen_knowledge_2022} & 78.59 $\pm$ 0.31 & 78.59 $\pm$ 0.05\\
\hline
ATSC & \textbf{79.38 $\pm$ 0.13} & \textbf{79.21 $\pm$ 0.12}\\
\specialrule{2pt}{1pt}{1pt}	
\end{tabular}
\vspace{-5pt}
\end{table}		

\textbf{Results on ImageNet.} We evaluate the performances achieved by various state-of-the-art KD methods on the large-scale ImageNet dataset across different numbers of training epochs to compare their convergence speeds and postconvergence performances. As detailed in Table \ref{Table_4}, the ATSC method not only converges faster than the other methods but also achieves the highest top-1 accuracy upon convergence, demonstrating its superior effectiveness for use with large-scale datasets.

\begin{table}[h]
\scriptsize
\centering
\caption{Comparison among the top-1 test accuracies (\%) achieved by various state-of-the-art KD methods on ImageNet with different numbers of training epochs. We adopt ResNet-50 as the teacher model and ResNet-18 as the student model. The pretrained teacher model for ATSC achieves a top-1 accuracy of 76.25\%, whereas the teacher model for the other methods reaches 76.26\%, as reported in \cite{chen_knowledge_2022}.}
\label{Table_4}
\begin{tabular}{c|ccccccccc|c}
\specialrule{2pt}{1pt}{1pt}								
Epochs & Student & KD~\cite{hinton_distilling_2014} & AT~\cite{zagoruyko_paying_2017} & SP~\cite{tung_similarity-preserving_2019-1} & VID~\cite{ahn_variational_2019} & CRD~\cite{tian_contrastive_2020} & SRRL~\cite{yang_knowledge_2021} & SemCKD~\cite{chen_cross-layer_2021} & SimKD~\cite{chen_knowledge_2022} & ATSC\\
\hline
30 & 49.34 & 52.75 & 52.85 & 53.57 & 53.22 & 55.44 & 55.14 & 53.14 & 61.73 & 66.67\\
60 & 64.98 & 66.69 & 66.69 & 66.36 & 66.64 & 67.25 & 67.36 & 66.89 & 69.26 & 70.84\\
120 & 70.58 & 71.29 & 71.18 & 71.08 & 71.11 & 71.25 & 71.46 & 71.41 & 71.66 & 71.77\\
\specialrule{2pt}{1pt}{1pt}	
\end{tabular}
\end{table}	

\subsection{Ablation study}\label{sec:ablation}
\textbf{The effect of the classifier used by the fine-tuning teacher network.} During the learning process of ATSC, the shared classifier is updated according to (\ref{eq_step2}) using the encoder of the teacher after updating the encoders of both the teacher and the student, as specified in (\ref{eq_step1}). However, without loss of generality, the classifier could alternatively be updated based on the encoder of the student, as illustrated in the following equation:
\begin{equation} \label{eq_step2_temp}
\min_{\bm{\theta}_C} \mathcal{L}_{CE}(\bm{y}, \sigma(C(\mathcal{P}(E_S(\bm{x}))))).
\end{equation}
This modification aims to directly align the classifier with the representations of the student model. We compare teacher-based fine-tuning, as specified in (\ref{eq_step2}), and student-based fine-tuning, as specified in (\ref{eq_step2_temp}).

As presented in Table \ref{Table_5}, our approach with teacher-based fine-tuning achieves a 0.43\% accuracy improvement over student-based fine-tuning, which directly learns to map its own representations to the ground-truth labels. This outcome demonstrates that leveraging the advanced classifier of a powerful teacher can enhance the performance of a student more effectively than developing a student-specific classifier.

% \begin{table}[h]
% \scriptsize
% \centering
% \caption{Comparison of teacher-based fine-tuning and student-based fine-tuning in terms of top-1 test accuracy (\%) for the ResNet-32x4 \& WRN-16-2 and ResNet-32x4 \& MobileNetV2x2 scenarios. The average and standard deviation are calculated from separate 4 trials.}
% \label{Table_5}
% \begin{tabular}{l|cc}
% \specialrule{2pt}{1pt}{1pt}								
% Teacher & ResNet-32x4 & ResNet-32x4\\
% Student & WRN-16-2 & MobileNetV2x2\\
% \hline
% Student-based & 76.58 $\pm$ 1.03 & 76.09 $\pm$ 0.13\\
% Teacher-based (Ours) & \textbf{77.34 $\pm$ 0.18} & \textbf{76.18 $\pm$ 0.16}\\
% \specialrule{2pt}{1pt}{1pt}	
% \end{tabular}
% \end{table}		

\textbf{Comparison with online baselines.} To evaluate the effectiveness of using a pretrained teacher in ATSC, we compare it with the following three baselines.
\begin{itemize}[leftmargin=1em]
\vspace{-8pt}
\item SimKD \cite{chen_knowledge_2022}: This baseline utilizes offline feature distillation while retaining the classifier from its pretrained teacher network.
\vspace{-3pt}
\item Online SimKD (O-SimKD): An online variant of SimKD, this approach begins with an initialized teacher network. For each batch, the parameters of the student network and the projector are updated after updating the teacher. %The objective function used is $\min_{\bm{\theta}_{E_S}, \bm{\theta}_{\mathcal{P}}} \mathcal{L}_{MSE}(E_T(\bm{x}), \mathcal{P}(E_S(\bm{x})))+\alpha * \mathcal{L}_{MSE}(\bm{\theta}^*_{E_T}, \bm{\theta}_{E_T})$.
\vspace{-3pt}
\item Online ATSC (O-ATSC): This baseline also starts with an initialized teacher network. Specifically, updates to the teacher network, (\ref{eq_step1}), and (\ref{eq_step2}) are sequentially applied for each batch.
\vspace{-8pt}
\end{itemize}
The experimental setups described in Table \ref{Table_5} are also used for this comparison.

Table \ref{Table_6} demonstrates that applying ATSC with an initialized teacher in an online manner results in a performance that is inferior to those of both SimKD and O-SimKD. However, starting the training process with pretrained teachers leads to accuracy improvements of 1.43\% and 1.09\% in the ResNet-32x4 \& WRN-16-2 and ResNet-32x4 \& MobileNetV2x2 scenarios, respectively. These results showcase the best performance achieved in each scenario and underscore the significant contribution of the discriminative power provided by the pretrained teacher in ATSC.

% \begin{table}[h]
% \scriptsize
% \centering
% \caption{Top-1 test accuracy (\%) comparison of baselines to evaluate the effectiveness of using a pretrained teacher.}
% \label{Table_6}
% \begin{tabular}{l|cc}
% \specialrule{2pt}{1pt}{1pt}								
% Teacher & ResNet-32x4 & ResNet-32x4\\
% Student & WRN-16-2 & MobileNetV2x2\\
% \hline
% SimKD & 77.17 $\pm$ 0.32 & 75.43 $\pm$ 0.26\\
% O-SimKD & 77.09 $\pm$ 0.19 & 75.98 $\pm$ 0.21\\
% O-ATSC & 75.91 $\pm$ 0.23 & 75.09 $\pm$ 0.20\\
% ATSC & \textbf{77.34 $\pm$ 0.18} & \textbf{76.18 $\pm$ 0.16}\\
% \specialrule{2pt}{1pt}{1pt}	
% \end{tabular}
% \end{table}		

\begin{table}[h]
\centering
\begin{minipage}[t]{0.57\textwidth}
\scriptsize
\centering
\caption{Comparison between teacher-based fine-tuning and student-based fine-tuning in terms of the top-1 test accuracies (\%) achieved in the ResNet-32x4 \& WRN-16-2 and ResNet-32x4 \& MobileNetV2x2 scenarios. The average and standard deviation values are calculated from separate 4 trials.}
\label{Table_5}
\begin{tabular}{l|cc}
\specialrule{2pt}{1pt}{1pt}								
Teacher & ResNet-32x4 & ResNet-32x4\\
Student & WRN-16-2 & MobileNetV2x2\\
\hline
Student-based & 76.58 $\pm$ 1.03 & 76.09 $\pm$ 0.13\\
Teacher-based (Ours) & \textbf{77.34 $\pm$ 0.18} & \textbf{76.18 $\pm$ 0.16}\\
\specialrule{2pt}{1pt}{1pt}	
\end{tabular}
\end{minipage}
\hfill
\begin{minipage}[t]{0.38\textwidth}
\scriptsize
\centering
\caption{Top-1 test accuracy (\%) comparison between the baselines and the proposed approach for evaluating the effectiveness of using a pretrained teacher.}
\label{Table_6}
\begin{tabular}{l|cc}
\specialrule{2pt}{1pt}{1pt}								
Teacher & ResNet-32x4 & ResNet-32x4\\
Student & WRN-16-2 & MobileNetV2x2\\
\hline
SimKD & 77.17 $\pm$ 0.32 & 75.43 $\pm$ 0.26\\
O-SimKD & 77.09 $\pm$ 0.19 & 75.98 $\pm$ 0.21\\
O-ATSC & 75.91 $\pm$ 0.23 & 75.09 $\pm$ 0.20\\
ATSC & \textbf{77.34 $\pm$ 0.18} & \textbf{76.18 $\pm$ 0.16}\\
\specialrule{2pt}{1pt}{1pt}	
\end{tabular}
\end{minipage}
\end{table}

\begin{figure}[h]
\centering
\begin{minipage}[t]{0.4\textwidth}
\scriptsize
\centering
\captionof{table}{Top-1 test accuracy (\%) changes exhibited after adapting pretrained teachers. `Teacher' denotes the performance of the pretrained teacher. We report the average accuracy ($\pm$ standard deviation) of the teacher model after training through ATSC over 4 trials.}
\label{Table_7}
\vspace{0pt}
\begin{tabular}{c|cc}
\specialrule{2pt}{1pt}{1pt}								
Teacher & ResNet-32x4 & ResNet-32x4\\
Student & WRN-16-2 & MobileNetV2x2\\
\hline
Teacher & 79.32 & 79.32\\
Adapted teacher & 79.03 $\pm$ 0.21 & 78.97 $\pm$ 0.10\\
\specialrule{2pt}{1pt}{1pt}	
\end{tabular}
\end{minipage}
\hfill
\begin{minipage}[t]{0.56\textwidth}
\vspace{0pt}
% \vspace{-10pt}
\centering
\includegraphics[width=0.7\linewidth]{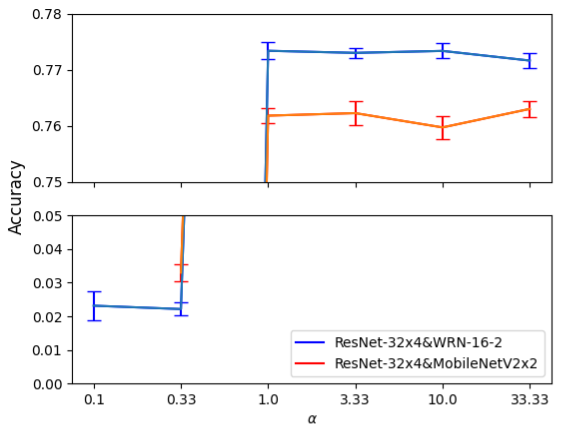}
\caption{Top-1 mean accuracies (with standard deviations) achieved over 4 separate trials under different values of the balancing parameter $\alpha$.}
\label{fig_sens}
% \vspace{-30pt}
\end{minipage}
\vspace{-10pt}
\end{figure}

\textbf{Changes in the performance of the teacher model.} Since the teacher model adapts its parameters to aid the student model during the ATSC training process, it may sacrifice some discriminative power. Consequently, we measure this change as an accuracy loss from the perspective of the teacher, as detailed in Table \ref{Table_7}. Overall, an accuracy loss of 0.32\% is observed. Generally, during KD, high-performance teachers are more likely to transfer effective learning patterns, thereby producing high-performance students. However, a comparison between the performances of SimKD and ATSC suggests that students can achieve more significant learning gains if the teacher provides more effective guidance, even at the cost of slightly diminished capabilities for the teacher. This is supported by the fact that other existing methods attain inferior performance in the ResNet-32x4 \& WRN-16-2 and ResNet-32x4 \& MobileNetV2x2 scenarios, as shown in Table \ref{Table_1}.

% \begin{table}[h]
% \scriptsize
% \centering
% \caption{Top-1 test accuracy (\%) changes from adapting pretrained teachers. `Teacher' denotes the performance of the pretrained teacher. We report the average accuracy ($\pm$ standard deviation) of the teacher model after training through ATSC over 4 trials.}
% \label{Table_7}
% \begin{tabular}{c|cc}
% \specialrule{2pt}{1pt}{1pt}								
% Teacher & ResNet-32x4 & ResNet-32x4\\
% Student & WRN-16-2 & MobileNetV2x2\\
% \hline
% Teacher & 79.32 & 79.32\\
% Adapted teacher & 79.03 $\pm$ 0.21 & 78.97 $\pm$ 0.10\\
% \specialrule{2pt}{1pt}{1pt}	
% \end{tabular}
% \end{table}		

\subsection{Sensitivity analysis}
Fig. \ref{fig_sens} presents the results of a sensitivity analysis conducted on the balancing parameter $\alpha$ within the ResNet-32x4 \& WRN-16-2 and ResNet-32x4 \& MobileNetV2x2 scenarios. We discover that when $\alpha$ exceeds 33.33, specifically reaching 100.00, the training loss diverges. Consequently, we set 33.33 as the maximum value for this analysis. The results indicate that ATSC is relatively robust to variations in $\alpha$, provided that it is neither too small nor too large (specifically, when it ranges between 1 and 33.33 in Fig. \ref{fig_sens}), which reduces the effort required for hyperparameter optimization. We also verify the performance changes produced according to different reduction factors, and the results are provided in the Appendix.

% \begin{wrapfigure}{r}{0.5\textwidth}
% \begin{center}
% \vspace{-10pt}
% \centerline{\includegraphics[width=\linewidth]{sensitivity_analysis.png}}
% \caption{Top-1 mean accuracy with standard deviation over 4 separate trials against the balancing parameter $\alpha$.}
% \label{fig_sens}
% \vspace{-30pt}
% \end{center}
% \end{wrapfigure}

\section{Discussion}\label{sec:conclusion}
In this paper, we introduce ATSC, a novel KD method grounded in three key insights: the robust discriminative power of a large-scale, pretrained teacher model; the self-adaptive teaching ability of this teacher during the training process; and the benefits of sharing the advanced classifier of the teacher. This study introduces the concept of adaptive teaching to the KD field for the first time, and we empirically demonstrate its significant contributions to the learning process. We conduct extensive experiments, and the results indicate that our proposed ATSC method substantially outperforms other state-of-the-art KD methods in both single-teacher and multiteacher scenarios. Furthermore, ATSC exhibits robustness across a wide range of balancing parameter settings, thereby simplifying the hyperparameter optimization process.

\subsection{Limitations and future work} \label{sec:limit}
Although our method achieves enhanced classification performance with only a small increase in the number of parameters required for the projector, this addition may still impose a burden on devices with limited resources. Therefore, we plan to develop a projector-free architecture that is suitable for ATSC or an enhanced version of it. In this study, we have validated the effectiveness of ATSC solely for classification tasks. However, given that our method is readily adaptable to other applications, such as segmentation and object detection, and can be extended to fields such as natural language processing and speech recognition, we intend to further explore these possibilities.

\subsection{Broader impacts} \label{sec:broader}
This work aims to make a significant contribution to the field of KD. Although the social impact of our proposed method is challenging to forecast due to its universal applicability across diverse fields, the findings of this study are expected to have a positive influence on a wide range of applications.

\small
\bibliographystyle{abbrv}
\bibliography{references}\ %IEEEabrv instead of IEEEfull

%from here, it is appendix
\newpage
\appendix

% \section{Appendix / supplemental material}
\section{Appendix}

\subsection{Pseudocode for ATSC}
The proposed ATSC method implements (\ref{eq_step1}) to update the encoders of both the teacher and student models, as well as the projector, and it uses (\ref{eq_step2}) to align the classifier of the teacher with the updated encoder for each batch. Accordingly, ATSC is summarized in the following pseudocode.

\begin{algorithm}
\caption{Pseudocode for ATSC}
\label{alg_1}
\begin{algorithmic}[1]
\newcommand{\INPUT}{\item[\algorithmicinput]}
\newcommand{\algorithmicinput}{\textbf{Input:}}
\newcommand{\OUTPUT}{\item[\algorithmicoutput]}
\newcommand{\algorithmicoutput}{\textbf{Output:}}

\INPUT {A training dataset $\mathcal{D}$; A pretrained teacher network encoder $E_T$ and classifier $C$, an initialized student network encoder $E_S$, and a projector $\mathcal{P}$, whose parameter sets are $\bm{\theta}_{E_T}, \bm{\theta}_{C}, \bm{\theta}_{E_S}$ and $\bm{\theta}_{\mathcal{P}}$, respectively;}
\OUTPUT {A well-trained student network with a shared classifier $\sigma \circ C\circ \mathcal{P} \circ E_S$;}

\WHILE{the student has not converged}
\STATE Sample a minibatch $\mathcal{B}$ from $\mathcal{D}$;
\STATE Update $\bm{\theta}_{E_T}$, $\bm{\theta}_{E_S}$, and $\bm{\theta}_{\mathcal{P}}$ based on\\
 \hfill $\min_{\bm{\theta}_{E_T}, \bm{\theta}_{E_S}, \bm{\theta}_{\mathcal{P}}} \sum_{\bm{x} \in \mathcal{B}} \mathcal{L}_{MSE}(E_T(\bm{x}), \mathcal{P}(E_S(\bm{x})))+\alpha * \mathcal{L}_{MSE}(\bm{\theta}^*_{E_T}, \bm{\theta}_{E_T})$;  \hfill
\STATE Update $\bm{\theta}_{C}$ based on \\
\centering
$\min_{\bm{\theta}_{C}} \sum_{\bm{x} \in \mathcal{B}} \mathcal{L}_{CE}(Y, \sigma(C(E_T(\bm{x}))))$;
\ENDWHILE
\end{algorithmic}
\end{algorithm}

\subsection{Projector}
To allow the student model to utilize the classifier of a teacher model with a different structure, an additional projector is necessary to align the student encoder with the classifier of the teacher. The structure of this projector is summarized in Table \ref{Table_projector}. The projector consists of three convolutional layers, each of which is followed by standard batch normalization and rectified linear unit (ReLU) activation. It is assumed that the spatial dimensions of the feature maps (with the height and width denoted as H and W, respectively) in both the teacher and student networks are the same. If the feature map of the teacher is larger, we apply an average pooling operation to the encoder of the teacher beforehand to align the spatial dimensions, reducing the imposed computational demand.

In the projector, the number of filters in each layer, and thereby the total number of parameters, is controlled by a single hyperparameter $r$. A lower $r$ value allows the student to learn more extensively, generally resulting in higher performance. However, a lower $r$ value also leads to a heavier projector, increasing the total number of parameters required by the student model, which may not always be desirable. Therefore, we report the performance achieved by the learned student model with different reduction factors in our experimental results. The changes in the number of parameters are detailed in Section \ref{sec:append_exp_result} of the Appendix.

\begin{table}[h]
\scriptsize
\centering
\caption{Summary of the projector architecture. Conv($x$,$x$) denotes an $x\times x$ convolution layer, and $r$ is the reduction factor that controls the number of filters contained in each layer.}
\label{Table_projector}
\begin{tabular}{c|c|c|c}
\specialrule{2pt}{1pt}{1pt}								
&Input & Operator & Output\\
\hline
1st layer&$H \times W \times Ch_S$& Conv(1,1) & $H \times W \times Ch_T/r$\\
2nd layer&$H \times W \times Ch_T/r$& Conv(3,3) & $H \times W \times Ch_T/r$\\
3rd layer&$H \times W \times Ch_T/r$& Conv(1,1) & $H \times W \times Ch_T$\\
\specialrule{2pt}{1pt}{1pt}	
\end{tabular}
\end{table}	

\subsection{Experimental settings} \label{sec:exp_setting_appen}

\textbf{Datasets.} Our experiments utilize two widely recognized image classification datasets: CIFAR-100 and ImageNet. The images acquired from each dataset are normalized based on their channel means and standard deviations. CIFAR-100 comprises 50,000 training images and 10,000 test images across 100 classes; each training image is padded by 4 pixels on all sides before being randomly cropped to a 32x32 format. Moreover, ImageNet consists of approximately 1.3 million training images and 50,000 validation images across 1,000 classes, where each image is randomly cropped to 224x224 without padding.

\textbf{Network structures.} We evaluate the performance achieved when using a wide array of teacher-student combinations; these include several widely used neural network architectures: VGG \cite{simonyan_very_2015}, ResNet \cite{he_deep_2016}, WRN \cite{zagoruyko_wide_2016}, MobileNetV2 \cite{sandler_mobilenetv2_2018}, ShuffleNetV1 \cite{zhang_shufflenet_2018}, and ShuffleNetV2 \cite{ma_shufflenet_2018}. The suffixes in networks labeled 'VGG-' and 'ResNet-' indicate the depths of the respective networks. For 'WRN-d-w', 'd' represents the depth, and 'w' indicates the width factor of the wide-ResNet. Following previous works \cite{tian_contrastive_2020, chen_cross-layer_2021, chen_knowledge_2022}, we adjust the number of convolution filters contained in the intermediate layers of some architectures by a specific ratio, which is denoted as 'x' in their names. For example, the notation 'ResNet-32x4' specifies a ResNet architecture that is 32 layers deep and whose convolution filters are expanded by a factor of four.

\textbf{Comparison methods.} In this paper, we compare various offline and online KD methods. A summary of the KD methods used for the comparison is provided in Table \ref{Table_KD_summary}. A more detailed summary of these key KD methods is available in Section \ref{sec:RW}. We categorize the methods as online approaches if they adjust the parameters of their teachers during the training phase.

\begin{table}[h]
\scriptsize
\centering
\caption{Summary of the compared KD methods}
\label{Table_KD_summary}
\begin{tabular}{l|l|l|l}
\specialrule{2pt}{1pt}{1pt}								
Abbreviation & Method & Mode & Reference\\
\hline
KD&Vanilla Knowledge Distillation&Offline&\cite{hinton_distilling_2014}\\
FitNet&FitNet&Offline&\cite{romero_fitnets_2015}\\
SP&Similarity-Preserving knowledge distillation&Offline&\cite{tung_similarity-preserving_2019-1}\\
RKD&Relational Knowledge Distillation&Offline&\cite{park_relational_2019}\\
CRD&Contrastive Representation Distillation&Offline&\cite{tian_contrastive_2020}\\
SRRL&Softmax Regression Representation Learning&Offline&\cite{yang_knowledge_2021}\\
ReviewKD&Knowledge Distillation via knowledge Review&Offline&\cite{chen_distilling_2021}\\
SimKD&Simple Knowledge Distillation&Offline&\cite{chen_knowledge_2022}\\
T$f$-FD&Teacher free Feature Distillation&Online&\cite{li_self-regulated_2022}\\
ONE&On-the-fly Native Ensemble&Online&\cite{lan_knowledge_2018}\\
KDCL&Online Knowledge Distillation via Collaborative Learning&Online&\cite{guo_online_2020}\\
MetaDistil&Knowledge Distillation with Meta Learning&Online&\cite{zhou_bert_2022}\\
DML&Deep Mutual Learning&Online&\cite{zhang_deep_2018}\\
SHAKE&SHAdow KnowlEdge transfer framework&Online&\cite{li_shadow_2022}\\
NORM&N-to-One Representation Matching&Offline&\cite{liu_norm_2023}\\
AT&Attention Transfer&Offline&\cite{zagoruyko_paying_2017}\\
VID&Variational Information Distillation&Offline&\cite{ahn_variational_2019}\\
SemCKD&Semantic calibration for Crosslayer Knowledge Distillation&Offline&\cite{chen_cross-layer_2021}\\
DistPro&Fast knowledge Distillation Process via meta optimization&Offline&\cite{deng_distpro_2022}\\
\specialrule{2pt}{1pt}{1pt}	
\end{tabular}
\end{table}	

\subsection{Experimental results} \label{sec:append_exp_result}
In our method, the reduction factor $r$ can significantly impact the resulting performance. Therefore, we evaluate the performance achieved not only with the default $r$ value of 2 but also with an $r$ value of 1. We additionally report the performance attained for the ResNet-50 \& VGG-8 scenario with an $r$ value of 4, as lower reduction factors of 1 and 2 do not substantially increase the parameter counts required in other scenarios, thus eliminating the need for a higher $r$, which typically degrades performance. The top-1 test accuracy results produced across diverse teacher \& student scenarios are summarized in Tables \ref{Table_1_appen} and \ref{Table_2_appen}.

Since our method involves adding a projector to the student model with a shared classifier obtained from the teacher, the parameter count of the student model varies with $r$. The numbers of network parameters required for the teacher model, the student model without a projector, and the student model with a projector are detailed in Tables \ref{Table_1_appen_para} and \ref{Table_2_appen_para}.

A comparison between the results obtained with an $r$ of 1 and an $r$ of 2 reveal that a lower $r$ value yields greater performance. Specifically, setting $r$ to 1 results in a 0.38\% accuracy gain over that attained within an $r$ of 2 but at the cost of a 21.04\% increase in the required number of parameters, which represents a significant tradeoff in terms of parameter efficiency. Therefore, we recommend selecting an $r$ value that balances performance with the required parameter count. In this paper, $r$ is set to 2 by default, with exceptions in a few specific cases.

\begin{table}[h]
\scriptsize
\centering
\caption{Top-1 test accuracies (\%) achieved by our ATSC method on the CIFAR-100 dataset with reduction factors of 1 and 2 in the teacher-student scenarios described in Table \ref{Table_1}. The results show the mean accuracy ± standard deviation values attained based on 4 runs.}
\label{Table_1_appen}
\begin{tabular}{c|c|cccccc}
\specialrule{2pt}{1pt}{1pt}
\multicolumn{2}{c|}{Teacher} & ResNet-32x4 & ResNet-32x4 & ResNet-32x4 & ResNet-32x4 & WRN-40-2 & ResNet-32x4\\
\multicolumn{2}{c|}{Student} & VGG-8 & ShuffleNetV2 & ShuffleNetV1 & WRN-16-2 & MobileNetV2 & MobileNetV2x2 \\
\hline
\multicolumn{2}{c|}{Teacher} & 79.32 & 79.32 & 79.32 & 79.32 & 76.44 & 79.32\\
\multicolumn{2}{c|}{Student} & 70.46$\pm$0.29 & 72.60$\pm$0.12 & 71.36$\pm$0.25 & 73.51$\pm$0.32 & 65.43$\pm$0.29 & 69.06$\pm$0.10 \\
\hline
Reduction&1&76.47$\pm$0.17&78.99$\pm$0.15&77.91$\pm$0.06&77.87$\pm$0.15&71.56$\pm$0.18&76.08$\pm$0.23\\
\cline{2-8}
factor&2&76.31$\pm$0.39&78.84$\pm$0.13&77.76$\pm$0.08&77.34$\pm$0.15&71.18$\pm$0.33&76.18$\pm$0.14\\
\specialrule{2pt}{1pt}{1pt}	
\end{tabular}
\end{table}

\begin{table}[h]
\scriptsize
\centering
\caption{Top-1 test accuracies (\%) achieved by our ATSC method on the CIFAR-100 dataset with reduction factors of 1 and 2 in the teacher-student scenarios described in Table \ref{Table_2}. The results show the mean accuracy ± standard deviation values attained based on 5 runs.}
\label{Table_2_appen}
\begin{tabular}{c|c|cccc}
\specialrule{2pt}{1pt}{1pt}								
\multicolumn{2}{c|}{Teacher} & ResNet-110 & ResNet-32x4 & VGG-13 & ResNet-50\\
\multicolumn{2}{c|}{Student} & ResNet-20 & ResNet-8x4 & MobileNetV2 & VGG-8\\
\hline
\multicolumn{2}{c|}{Teacher} & 73.88 & 79.32 & 74.85 & 78.87\\
\multicolumn{2}{c|}{Student} & 69.06 & 72.50 & 64.60 & 70.36\\
\hline
\multirow{3}{*}{\shortstack{Reduction \\ factor}}&1 & 69.28$\pm$0.30 & 78.80$\pm$0.17 & 70.66$\pm$0.34 & 76.71$\pm$0.19\\
\cline{2-6}
&2 & 67.42$\pm$0.21 & 78.27$\pm$0.12 & 70.21$\pm$0.22 & 77.02$\pm$0.16\\
\cline{2-6}
&4 & N/A & N/A & N/A & 76.53$\pm$0.21\\
\specialrule{2pt}{1pt}{1pt}	
\end{tabular}
\end{table}

\begin{table}[h] %SimKD + Norm
\scriptsize
\centering
\caption{Numbers of parameters required by the teacher model, the student model without a projector, and the student model with a projector across various reduction factors for the scenarios described in Table \ref{Table_1}. The ratios of the parameter increases induced by adding the projector to the student model, relative to the numbers of teacher model parameters, are also reported in parentheses.}
\label{Table_1_appen_para}
\begin{tabular}{c|c|cccccc}
\specialrule{2pt}{1pt}{1pt}
\multicolumn{2}{c|}{Teacher} & ResNet-32x4 & ResNet-32x4 & ResNet-32x4 & ResNet-32x4 & WRN-40-2 & ResNet-32x4\\
\multicolumn{2}{c|}{Student} & VGG-8 & ShuffleNetV2 & ShuffleNetV1 & WRN-16-2 & MobileNetV2 & MobileNetV2x2 \\
\hline
\multicolumn{2}{c|}{Teacher} & 7.4 M & 7.4 M & 7.4 M & 7.4 M & 2.3 M & 7.4 M\\
\multicolumn{2}{c|}{Student (w/o projector)} & 4.0 M & 1.4 M & 0.9 M & 0.7 M & 0.8 M & 2.4 M\\
\hline
\multirow{4}{*}{\shortstack{Reduction \\ factor}}&\multirow{2}{*}{1}&4.8 M & 2.3 M & 1.9 M & 1.4 M & 1.2 M & 3.4 M\\
&& (7.28\%) & (8.16\%) & (8.05\%) & (6.62\%) & (8.91\%) & (8.60\%)\\
\cline{2-8}
&\multirow{2}{*}{2}&4.2 M & 1.7 M & 1.3 M & 0.9 M & 1.0 M & 2.7 M\\
&& (3.66\%) & (4.55\%) & (4.44\%) & (3.00\%) & (6.23\%) & (4.99\%)\\
\specialrule{2pt}{1pt}{1pt}	
\end{tabular}
\end{table}

\begin{table}[h]
\scriptsize
\centering
\caption{Numbers of parameters required by the teacher model, the student model without a projector, and the student model with a projector across various reduction factors for the scenarios described in Table \ref{Table_2}. }
\label{Table_2_appen_para}
\begin{tabular}{c|c|cccc}
\specialrule{2pt}{1pt}{1pt}								
\multicolumn{2}{c|}{Teacher} & ResNet-110 & ResNet-32x4 & VGG-13 & ResNet-50\\
\multicolumn{2}{c|}{Student} & ResNet-20 & ResNet-8x4 & MobileNetV2 & VGG-8\\
\hline
\multicolumn{2}{c|}{Teacher} & 1.7 M & 7.4 M & 9.5 M & 23.7 M\\
\multicolumn{2}{c|}{Student (w/o projector)} & 0.3 M & 1.2 M & 0.8 M & 4.0 M \\
\hline
\multirow{6}{*}{\shortstack{Reduction \\ factor}}&\multirow{2}{*}{1} & 0.3 M & 2.0 M & 4.1 M & 47.2 M\\
&& (2.99\%) & (10.06\%) & (35.21\%) & (182.28\%)\\
\cline{2-6}
&\multirow{2}{*}{2} & 0.3 M & 1.5 M & 1.9 M & 16.2 M\\
&& (1.16\%) & (3.22\%) & (11.65\%) & (51.77\%)\\
\cline{2-6}
&\multirow{2}{*}{4} & \multirow{2}{*}{N/A} & \multirow{2}{*}{N/A} & \multirow{2}{*}{N/A} & 7.8 M\\
&& &  &  & (16.37\%)\\
% &2 & 67.42$\pm$0.21 & 78.27$\pm$0.12 & 70.21$\pm$0.22 & 77.02$\pm$0.16\\
% &4 & N/A & N/A & N/A & 76.53$\pm$0.21\\
\specialrule{2pt}{1pt}{1pt}	
\end{tabular}
\end{table}		

%%%%%%%%%%%%%%%%%%%%%%%%%%%%%%%%%%%%%%%%%%%%%%%%%%%%%%%%%%%%

\end{document}